\documentclass[sigconf,nonacm]{acmart}

\usepackage{multirow}
\usepackage{graphicx}
\usepackage{subcaption}
\usepackage{arydshln}
\usepackage{float}

\begin{document}

\title{On the Unreasonable Effectiveness of Centroids in Image Retrieval}

\author{Mikolaj Wieczorek}
\authornote{\label{equalContr}Both authors contributed equally to this research.}
\affiliation{%
  \institution{Synerise}
  \institution{Warsaw University of Technology}
  \country{Poland}
}

\author{Barbara Rychalska\footnotemark[1]}
\affiliation{%
  \institution{Synerise}
  \institution{Warsaw University of Technology}
  \country{Poland}
}

\author{Jacek Dabrowski}
\affiliation{%
  \institution{Synerise}
  \country{Poland}
}

\renewcommand{\shortauthors}{M. Wieczorek et al.}

\begin{abstract}
Image retrieval task consists of finding similar images to a query image from a set of gallery (database) images. Such systems are used in various applications e.g. person re-identification (ReID) or visual product search. 
Despite active development of retrieval models it still remains a challenging task mainly due to large intra-class variance caused by changes in view angle, lighting, background clutter or occlusion, while inter-class variance may be relatively low.
A large portion of current research focuses on creating more robust features and modifying objective functions, usually based on Triplet Loss. Some works experiment with using centroid/proxy representation of a class to alleviate problems with computing speed and hard samples mining used with Triplet Loss. However, these approaches are used for training alone and discarded during the retrieval stage. In this paper we propose to use the mean centroid representation both during training and retrieval. Such an aggregated representation is more robust to outliers and assures more stable features. As each class is represented by a single embedding - the class centroid - both retrieval time and storage requirements are reduced significantly. Aggregating multiple embeddings results in a significant reduction of the search space due to lowering the number of candidate target vectors, which makes the method especially suitable for production deployments. Comprehensive experiments conducted on two ReID and Fashion Retrieval datasets demonstrate effectiveness of our method, which outperforms the current state-of-the-art. We propose centroid training and retrieval as a viable method for both Fashion Retrieval and ReID applications.
\end{abstract}


\maketitle

\section{Introduction}

Instance retrieval is a problem of matching an object from a query image to objects represented by images from a gallery set. Applications of retrieval systems span person/vehicle re-identification,  face recognition, video surveillance, explicit content filtering, medical diagnosis and fashion retrieval.  

Most existing instance retrieval solutions use Deep Metric Learning methodology \cite{Diao2021_similarity_reasoning_dml, Jun_combination_dml, Wieczorek2020, luo_reid_strong_baseline, Liu_deepfashion, Yuan_in_defense_of_the}, in which a deep learning model is trained to transform images to a vector representation, so that samples from the same class are close to each other. At the retrieval stage, the query embedding is scored against all gallery embeddings and the most similar ones are returned.
Until recently, a lot of works used classification loss for the training of retrieval models \cite{Noh2017_class_loss, Xiao2016_class_loss, Zhai2018_class_loss, Zhou2019_class_loss, Xiao2016a_class_loss}.
Currently most works use comparative/ranking losses and the Triplet Loss is one of the most widely used approaches. However, state-of-the-art solutions often combine a comparative loss with auxiliary losses such as classification or center loss \cite{Wieczorek2020, luo_reid_strong_baseline, Wen2016_a_discriminative_feature, Lagunes-Fortiz_centroids_triplet, Yuan_in_defense_of_the}.

\begin{figure}[]
    \begin{subfigure}{0.52\columnwidth}
      \centering
      \includegraphics[width=1.\linewidth]{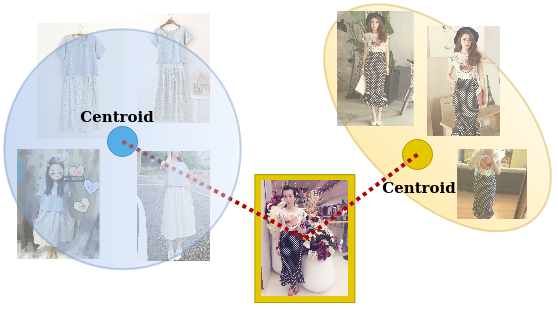}
      \caption{Centroid-based retrieval}
    \end{subfigure}
    \begin{subfigure}{0.46\columnwidth}
      \centering
      \includegraphics[width=1.\linewidth]{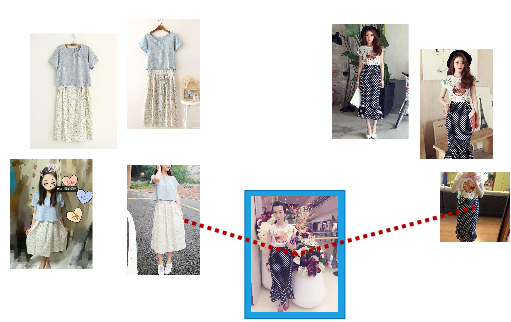}
      \caption{Instance-based retrieval}
    \end{subfigure}
    \caption{Comparison of centroid-based and  instance-based retrieval. Dashed lines indicate distance between the query image (coloured frame) and the nearest neighbour from each class. a) The centroid  is calculated as the mean of all samples (shaded images) belonging to each class. The query is assigned the class of the nearest centroid, which is the correct "gold" class. b) The distance is calculated between all samples and the query. It is erroneously assigned the "blue" class, as the blue-class sample is its nearest neighbour.}
    \label{fig:teaser_image}
\end{figure}

Even though Triplet Loss is superior to most other approaches, it has problems that were indicated by numerous works \cite{Do2019_triplet_upperbound, Zhang_rethinking_classifcation, Yuan_in_defense_of_the, Zhou_point_to_set}:
1) Hard negative sampling is the dominant approach in creating training batches containing only informative triplets in a batch, but it may lead to bad local minima and prevent the model from achieving top performance \cite{Do2019_triplet_upperbound, Zhang_rethinking_classifcation}; 2) Hard negative sampling is computationally expensive, as the distance needs to be calculated between all samples in the batch \cite{Do2019_triplet_upperbound, Yuan_in_defense_of_the}; 3) Triplet Loss is prone to outliers and noisy labels due to hard negative sampling and the nature of point-to-point losses \cite{Yuan_in_defense_of_the, Zhang_rethinking_classifcation}

To alleviate problems stemming from the point-to-point nature of Triplet Loss, changes to point-to-set/point-to-centroid formulations were proposed, where the distances are measured between a sample and a prototype/centroid representing a class. Centroids are aggregations of each item's multiple representations. A centroid approach results in one embedding per item, decreasing both memory and storage requirements. There are a number of approaches investigating the prototype/centroid formulation and their main advantages are as follows:
1) Lower computational cost \cite{Do2019_triplet_upperbound, Yuan_in_defense_of_the}, of even linear complexity instead of cubic \cite{Do2019_triplet_upperbound}; 2) Higher robustness to outliers and noisy labels \cite{Yuan_in_defense_of_the, Zhang_rethinking_classifcation}; 3) Faster training \cite{Wang_centroid_based}; 4) Comparable or better performance than the standard point-to-point triplet loss \cite{Wang_centroid_based,Yuan_in_defense_of_the,Lagunes-Fortiz_centroids_triplet}.

\begin{figure*}[!h]
  \centering
  \includegraphics[width=\textwidth]{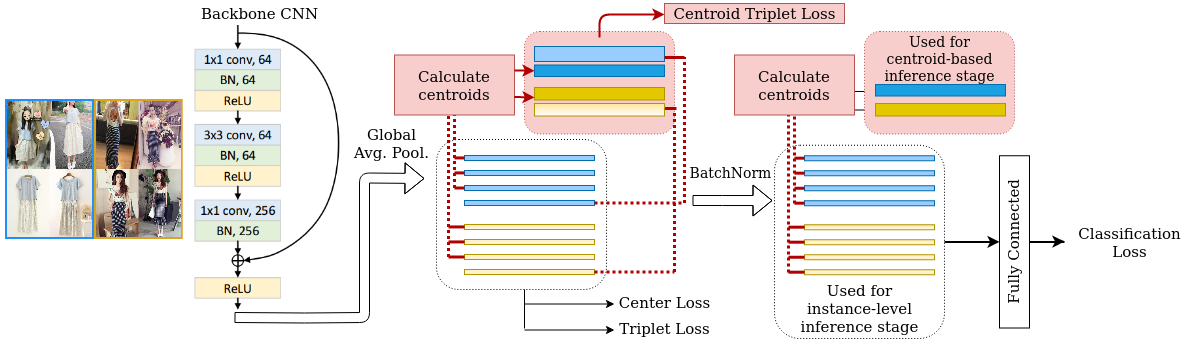}
  \caption{Architecture of our CTL-Model. Parts added over \cite{Wieczorek2020} are marked in red.}
  \Description{CTL Model architecture}
\label{fig:model}
\end{figure*}

We propose to go a step further and use the centroid-based approach for both training and inference, with applications to fashion retrieval and person re-identification. We implement our centroid-based model by augmenting the current state-of-the-art model in fashion retrieval  \cite{Wieczorek2020} with a new loss function we call Centroid Triplet Loss. The baseline model has a number of losses optimized simultaneously, which account for various aspects of the retrieval problem. An additional centroid-based loss can thus be easily added in order to amend one of the recurring problems: lack of robustness against variability in object galleries. Centroids are computed with simple averaging of image representations. We show that this straightforward model amendment allows to lower the latency of requests and decrease infrastructure costs, at the same time producing new state-of-the-art results in various evaluation protocols, datasets and domains.
We also discuss why such formulation of the retrieval problem is viable and advantageous compared to standard image-based approaches.

The contributions of this work are fourfold:
\begin{itemize}
    \item We introduce the Centroid Triplet Loss - a new loss function for instance retrieval tasks
    \item We propose to use class centroids as representations during retrieval.
    \item We show through thorough experiments that the centroid-based approach establishes new state-of-the-art results across different datasets and domains (fashion retrieval and person re-identification).
    \item We show that the centroid-based approach for retrieval tasks brings significant inference speed-ups and storage savings compared to the standard instance-level approach.
\end{itemize}

\section{Proposed method}

The image retrieval task aims to find the most similar object to the query image. In both fashion retrieval and person re-identification it is usually done on an instance-level basis: each query image is scored against all images from the gallery. If an object has several images assigned (e.g. photos from multiple viewpoints, under variable lighting conditions), then each image is treated separately. As a result, the same object may occur multiple times in the ranking result. Such a protocol can be beneficial as it allows to match images that were taken in similar circumstances, with similar angle, depicting the same part of the object or a close-up detail. On the other hand, the advantage can easily turn disadvantageous as a photo of a detail of a completely different object may be similar to the details in the query image, causing a false match.

We propose to use an aggregated item representation using all available samples. This approach results in a robust representation which is less susceptible to a single-image false matches. Using aggregated representations, each item is represented by a single embedding, leading to a significantly reduced search space - saving memory and reducing retrieval times significantly.
Apart from being more computationally efficient during retrieval, the centroid-based approach also improves retrieval results compared to non-centroid-based approaches.%
Note that training the model in a centroid-based setting does not restrict the evaluation protocol to centroid-only evaluation, but also improves results in the \textit{typical} setting of instance-level evaluation.

\subsection{Centroid Triplet Loss}

Triplet Loss originally works on an anchor image $A$, a positive (same class) example $P$ and a negative example belonging to another class $N$. The objective is to minimize the distance between $A-P$, while push away the $N$ sample. The loss function is formulated as follows:
\begin{equation}
    \mathcal{L}_{triplet} = \left[ \left\Vert f(A)-f(P) \right\Vert_2^2 - \left\Vert f(A)-f(N) \right\Vert_2^2 + \alpha  \right]_+\label{eq_1}
\end{equation}
where $[z]_+ = max(z,0)$, $f$ denotes embedding function learned during training stage and $\alpha$ is a margin parameter.

We propose the \textit{Centroid Triplet Loss (CTL)}. Instead of comparing the distance of an anchor image $A$ to positive and negative instances, CTL measures the distance between $A$ and class centroids $c_P$ and $c_N$ representing either the same class as the anchor or a different class respectively.
CTL is therefore formulated as:
\begin{equation}
    \mathcal{L}_{triplet} = \left[ \left\Vert f(A)-c_P \right\Vert_2^2 - \left\Vert f(A)-c_N) \right\Vert_2^2 + \alpha_c \right]_+\label{eq_2}
\end{equation}

\subsection{Aggregating item representations}

During training stage each mini-batch contains $P$ distinct item classes with $M$ samples per class, resulting in batch size of $P \times M$.
Let $S_k$ denote a set of samples for class $k$ in the mini-batch such that 
$\mathcal{S}_k = \{x_1, ..., x_M\}$ 
where $x_i$ represents an embedding of \emph{i-th} sample, such that $x_i \in R^D$, with $D$ being the sample representation size.
For effective training, each sample from $\mathcal{S}_k$ is used as a query $q_k$  and the rest $M-1$ samples are used to build a prototype centroid $c_{k_p}$, which can be expressed as:

\begin{equation}
  c_{k_p} = \frac{1}{\left|\mathcal{S}_k\setminus\{q_k\}\right|} \sum_{x_i \in \mathcal{S}_k\setminus\{q_k\}} f(x_i)
\end{equation}

where $f$ represents the neural network encoding images to $D$ dimensional embedding space.

During evaluation query images are supplied from the query set $\mathcal{Q}$, and centroids for each class $k$ are precalculated before the retrieval takes place. To construct these centroids we use all embeddings from the gallery set $\mathcal{G}_k$ for class $k$.
The centroid of each class $c_k \in R^D$ is calculated as the mean of all embeddings belonging to the given class: 

\begin{equation}
  c_k = \frac{1}{\left|\mathcal{G}_k\right|} \sum_{x_i \in \mathcal{G}_k} f(x_i)
\end{equation}

We apply centroid computation and CTL to the fashion retrieval state-of-the-art model described in \cite{Wieczorek2020}. This model embeds images with a baseline CNN model (using variations of the ResNet architecture) and passes them through a simple feed-forward architecture with average pooling and batch normalization. Three separate loss functions are computed at various stages of forward propagation. We add centroid computation for training just after embedding with the CNN. Centroids for inference are computed in the next step (after batch normalization) for consistency with the original model.  The resulting architecture is displayed in Figure \ref{fig:model}. Note that our centroid-based training and evaluation method can be also transplanted to other models, as CTL can be computed next to other existing loss functions.

\begin{table}[!ht]
\centering
\caption{Fashion Retrieval Results. S or L in the model name indicates input image size, either Small (256x128) or Large (320x320). R50 or R50IBN suffix indicates which backbone CNN was used, Resnet50 or Resnet50-IBN-A respectively. \text{'CE'} at the end of model name denotes Centroid-based Evaluation.}
\resizebox{\columnwidth}{!}{%
\begin{tabular}{c|l|ccccc}
{Dataset}       & \multicolumn{1}{c|}{Model}   & mAP            & Acc@1         & Acc@10         & Acc@20         & Acc@50      \\ \hline
\multirow{6}{*}{\rotatebox[origin=c]{90}{DeepFashion}}  &  SOTA (S-R50) \cite{Wieczorek2020}                                & 0.324          & 0.281         & 0.583          & 0.655          & 0.742               \\

                              & CTL-S-R50                                   & {0.344} & {0.298} & 
                        {0.612} & {0.685} & {0.770} \\ \cdashline{2-7} 
                        
                                                      &  CTL-S-R50 \textbf{CE}  &  \textbf{0.404} & \textbf{0.294} &  \textbf{0.613} & \textbf{0.689} & \textbf{0.774} \\ 
                              
                              \cline{2-7} 

                               & SOTA (L-R50IBN) \cite{Wieczorek2020}                              & 0.430          & \textbf{0.378} & 0.711          & 0.772          & 0.841 \\

                               & CTL-L-R50IBN                                & {0.431} & 0.376          & 
                        {0.711} & {0.776} & {0.847} \\ \cdashline{2-7}  
                        
                        &  CTL-L-R50IBN \textbf{CE} &  \textbf{0.492} & 0.373 &  \textbf{0.712} & \textbf{0.777} & \textbf{0.850} \\ 
                        
                        \hline
\multirow{6}{*}{\rotatebox[origin=c]{90}{Street2Shop}} & SOTA (S-R50) \cite{Wieczorek2020}                                 & 0.320          & 0.366    & {0.611} & 0.606          & -    \\

                              &  CTL-S-R50                                   & {0.353} & {0.418}          
                                     
                               & 0.594          & {0.643}          & 0.702
                        \\  \cdashline{2-7}     
                                                &  CTL-S-R50 \textbf{CE} &  \textbf{0.498}          & \textbf{0.432}      & \textbf{0.619}        & \textbf{0.660}        & \textbf{0.721}        \\ \cline{2-7} 

                              & SOTA (L-R50IBN) \cite{Wieczorek2020}                              & {0.468} & \textbf{0.537}     
                            & {0.698}
                               & {0.736} & -            \\
                               & CTL-L-R50IBN                                & 0.459          & 0.533   & 0.689          & 0.728          & 0.782
                               
                        \\ \cdashline{2-7}
                               &  CTL-L-R50IBN \textbf{CE}  & \textbf{0.598} & \textbf{0.537} &  \textbf{0.709} & \textbf{0.750} & \textbf{0.792}
\end{tabular}%
}
\label{tab:fashion-results}
\end{table}

\section{Experiments}

\subsection{Datasets}

\paragraph{DeepFashion (Fashion Retrieval).} The dataset was introduced by \cite{Liu_deepfashion} an  contains over 800,000 images, which are spread across several fashion related tasks. The data we used is a  \textit{Consumer-to-shop Clothes Retrieval} subset that contains 33,881 unique clothing products and 239,557 images. 

\paragraph{Street2Shop (Fashion Retrieval).} The dataset contains over 400,000 shop photos and 20,357 street photos. In total there are 204,795 distinct clothing items in the dataset. It is one of the first modern large-scale fashion dataset and was introduced by \cite{HadiKiapour_where_to_buy}. 

\paragraph{Market1501 (Person Re-identification).}
Introduced in \cite{Zheng2015_market1501} in 2015, it contains 1501 classes/ identities scattered across 32,668 bounding boxes and captured by 6 cameras at Tsinghua University. 751 classes are used for training, and 750 with distractors are used for evaluation.

\paragraph{DukeMTMC-reID (Person Re-identification).}
It is a subset of \\ DukeMTMC dataset \cite{Ristani2016_dukemtmcreid}. It contains 1,404 classes/identities, 702 are used for training and 702 along with 408 distractor identities are used for evaluation. 

\subsection{Implementation Details}
We implement our centroid-based solution on top of the current fashion retrieval state-of-the-art model \cite{Wieczorek2020}, which itself is based on a top-scoring ReID model \cite{luo_reid_strong_baseline}. We train our model on various Resnet-based backbones pretrained on ImageNet, and report results for Fashion Retrieval and Person Re-Identification tasks. We evaluate the model both in centroid-based and instance-based setting. Instance-based setting means that pairs of images are evaluated, identically as in the evaluation setting of \cite{Wieczorek2020}. We use the same training protocol presented in the aforementioned papers (e.g. random erasing augmentation, label smoothing), without introducing any additional steps.

\paragraph{Feature extractor}
We test two CNNs: Resnet-50 and Resnet50-IBN-A to compare our results on those two networks. Like \cite{Wieczorek2020, luo_reid_strong_baseline}, we use $stride=1$ for the last convolutional layer and Resnet-50 native 2048 dimensional embedding size.

\paragraph{Loss functions}
\cite{Wieczorek2020, luo_reid_strong_baseline} use a loss function consisting of three parts: (1) Triplet loss calculated on the raw embeddings, (2) Center Loss \cite{Wen2016_a_discriminative_feature} as an auxiliary loss, (3) classification loss computed on batch-normalized embeddings. To train our model based on centroids we use the same three losses and add CTL, which is computed between query vectors and class centroids. Center Loss was weighted by a factor of $5e^{-4}$, all other losses were assigned a weight of $1$.

Our Fashion Retrieval parameter configuration is
identical as in \cite{Wieczorek2020}. We use Adam optimizer with base learning rate of $1e^{-4}$ and multistep learning rate scheduler, decreasing the learning rate by a factor of 10 after 40\textsuperscript{th} and 70\textsuperscript{th} epoch. Like in \cite{Wieczorek2020, luo_reid_strong_baseline} the Center Loss was optimized separately by SGD optimizer with $lr=0.5$. Each model was trained 3 times, for 120 epochs each. 
For Person Re-Identification, the configuration is identical as in \cite{luo_reid_strong_baseline}. The  base learning rate is $3.5e^{-4}$, decayed at 40\textsuperscript{th} and 70\textsuperscript{th} epoch. The models were trained for 120 epochs each.

\paragraph{Resampling}

For Triplet Loss it is important to have enough positive samples per class, but some classes may have few samples. Therefore it is a common practice to define a target sample size $M$ and resample class instances if $\left|\mathcal{S}_k\right| < M$, resulting in repeated images in the mini-batch. We empirically verify that in our scenario it is beneficial to omit the resampling procedure. As resampling introduces noise to class centroids, we use only the unique class instances which are available.

\paragraph{Retrieval procedure} We follow \cite{luo_reid_strong_baseline} and \cite{Wieczorek2020}  in utilizing batch-normalized vectors during inference stage. Likewise, we use cosine similarity as the distance measure.
For the ReID datasets we use a cross-view matching setting, which is used in other ReID papers \cite{luo_reid_strong_baseline, Wang2019_reid_spatial_temporal}. This protocol ensures that for each query its gallery samples that were captured by the same camera are excluded during retrieval.

\begin{table}[h]
\centering
\caption{Comparison of storage and time requirements between instance and centroid-based models across tested datasets}
\label{tab:results_compute}
\resizebox{\columnwidth}{!}{%
\begin{tabular}{l|l|c|c|c}
\multicolumn{1}{c|}{Dataset}   & \multicolumn{1}{c|}{Mode} & \# in gallery & \begin{tabular}[c]{@{}c@{}}Embeddings\\ filesize (MB)\end{tabular} & \begin{tabular}[c]{@{}c@{}}Total eval \\ time (s)\end{tabular} \\ \hline
\multirow{2}{*}{Deep Fashion}  & Instances                    & 22k           & 175                                                                 & 81.35                                                          \\
                               & Centroids                 & 16k           & 130                                                                 & 59.83                                                          \\ \hline
\multirow{2}{*}{Street2Shop} & Instances                    & 350k          & 2700                                                                & 512.30                                                         \\
                               & Centroids                 & 190k          & 1500                                                                & 146.28                                                         \\ \hline
\multirow{2}{*}{Market1501}    & Instances                    & 16k           & 120                                                                 & 4.75                                                           \\
                               & Centroids                 & 0.75k         & 6                                                                   & 0.26                                                           \\ \hline
\multirow{2}{*}{Duke-MTMC}    & Instances                    & 17k           & 140                                                                 & 3.61                                                           \\
                               & Centroids                 & 1.1k          & 9                                                                   & 0.37                                                          
\end{tabular}%
}
\end{table}

\begin{table}[!h]
\centering
\caption{Person Re-Identification Results}
\resizebox{\columnwidth}{!}{%
\begin{tabular}{l|l|cccc}
Dataset                         & \multicolumn{1}{c|}{Model} & mAP            & Acc@1          & Acc@5          & Acc@10         \\ \hline
\multirow{2}{*}{Market1501}     & SOTA \cite{Wang2019_reid_spatial_temporal}                            & 0.955          & \textbf{0.980}           & \textbf{0.989}          & 0.991          \\

                                & CTL-S-R50                  & \textbf{0.983} & \textbf{0.980} & 0.986 & \textbf{0.995} \\ \hline
\multirow{2}{*}{Duke-MTMC-ReID} & SOTA  \cite{Wang2019_reid_spatial_temporal}                     & 0.927          & 0.945          & \textbf{0.968}          & 0.971          \\

                                & CTL-S-R50                  & \textbf{0.961} & \textbf{0.956} & 0.962 & \textbf{0.979}         
\end{tabular}%
}
\label{tab:reid-results}
\end{table}

\subsection{Fashion Retrieval Results}

We present the evaluation results for fashion retrieval in Table \ref{tab:fashion-results}. We evaluate two models: \texttt{SOTA} denotes the model presented in \cite{Wieczorek2020}, and \texttt{CTL} - our centroid-based model. Each model was evaluated in two modes: 1) standard instance-level evaluation on per-image basis (for both \texttt{SOTA} and \texttt{CTL} models), and 2) centroid-based evaluation, (denoted by \texttt{\textbf{CE}} in Table \ref{tab:fashion-results}): evaluation of \texttt{CTL} model on per-object basis, where all images from each class were used to build the class centorid and retrieval was done in centroid domain. 

Our CTL model performs better than the current state-of-the-art in most metrics across all tested datasets. Especially noticeable is the surge in mAP metric, which can be explained by the fact that usage of centroids reduces the search space. The reduction of the search space with centroid-based evaluation is coupled with reduction of the number of positive instances (from several to just one). Accuracy@K metrics on the other hand are not influenced by the change of search space.

\subsection{Person ReID Results}
We present the evaluation results for person re-identification in Table \ref{tab:reid-results}. Similarly as in fashion retrieval, we evaluate the following models: \texttt{SOTA} denotes the current state-of-the-art in ReID \cite{Wang2019_reid_spatial_temporal}, and \texttt{CTL} - our centroid-based model. We only report centroid-based evaluation results for CTL-model, as previous methods often restrict the search space arbitrarily.
For example, \cite{Wang2019_reid_spatial_temporal} (the current SOTA on both ReID test datasets) reduce the search space during retrieval with spatial and temporal constraints to decrease the number of candidates by eliminating cases where the person could not have possibly moved by a certain distance in the given time. Their approach requires extra information in the dataset and world knowledge necessary to construct the filtering rules, apart from just image understanding. Despite reliance on image matching alone, our centroid-based search space reduction achieves nearly the same or even better results across all metric on both datasets, outperforming \cite{Wang2019_reid_spatial_temporal} across most metrics and establishing the new state-of-the-art results.

\subsection{Memory Usage and Inference Times}

To test memory and computation efficiency of our centroid-based method compared to standard image-based retrieval, we compare the wall-clock time taken for evaluating all test datasets and the storage required for saving all embeddings.
Table \ref{tab:results_compute} shows the statistics for all datasets for instance-level and centroid-based scenarios. It can be seen that the centroid-based approach significantly reduces both retrieval time and the disk space required to store the embeddings. The reduction is caused by the fact that there are often several images per class, thus representing a whole group of object images with a centroid reduces the number of vectors necessary for a successful retrieval to one.

\section{Conclusions}
We introduce Centroid Triplet Loss - a new loss function for instance retrieval tasks. We empirically confirm that it significantly improves the accuracy of retrieval models. In addition to the new loss function, we propose the usage of class centroids during retrieval inference, further improving the accuracy metrics on retrieval tasks. Our methods are evaluated on four datasets from two different domains: Person Re-identification and Fashion Retrieval, and establish new state-of-the-art results on all datasets. In addition to accuracy improvements, we show that centroid-based inference leads to very significant computation speedups and lowering of memory requirements. The combination of increased accuracy with faster inference and lower resource requirements make our method especially useful in applied industrial settings for instance retrieval.

\bibliographystyle{ACM-Reference-Format}
\bibliography{bibliography}

\end{document}